\documentclass[english,american,conference, balance]{IEEEtran}
\usepackage[T1]{fontenc}
\usepackage[latin9]{inputenc}
\usepackage{array}
\usepackage{booktabs}
\usepackage{url}
\usepackage{multirow}
\usepackage{graphicx}

\makeatletter

\providecommand{\tabularnewline}{\\}

\usepackage{siunitx}
\usepackage{layouts}
\usepackage[table]{xcolor}
\usepackage{amsmath}
\definecolor{lightgray}{gray}{0.92}
\usepackage{graphicx}
\usepackage{listings}
\usepackage{xcolor}
\usepackage{url}
\usepackage{pifont}
\usepackage{multirow,bigdelim}
\usepackage{etoolbox}
\usepackage{hyperref}

\robustify\bfseries

\newcommand*\OK{\ding{51}}

\definecolor{codestrings}{rgb}{0.164,0,1}
\definecolor{codecomment}{rgb}{0.25,0.49,0.37}
\definecolor{codekeywords}{rgb}{0.8,0,0.33}
\definecolor{codebackground}{rgb}{0.95,0.95,0.95}
\lstdefinestyle{cblockstyle}{
inputencoding=utf8,
language=python,
extendedchars=true,
basicstyle=\ttfamily\footnotesize,
numbers=left,
  	numbersep=3pt,
framexleftmargin=2pt,
  	framerule=0pt,
  	frame=lines,
numberstyle=\tiny,
tabsize=2,
showstringspaces=false,
showspaces=false,
  keywordstyle=\bfseries\color{codekeywords},
  identifierstyle=\color{black},
  stringstyle=\color{codestrings},
  commentstyle=\color{codecomment},
  columns=fullflexible,
  abovecaptionskip=\medskipamount,
  belowcaptionskip=\medskipamount,
  backgroundcolor=\color{codebackground},
}
\lstdefinestyle{pblockstyle}{
inputencoding=utf8,
language=python,
extendedchars=true,
basicstyle=\ttfamily\footnotesize ,
numbers=left,
  	numbersep=3pt,
framexleftmargin=2pt,
  	framerule=0pt,
  	frame=lines,
numberstyle=\tiny,
tabsize=2,
showstringspaces=false,
showspaces=false,
  keywordstyle=\bfseries\color{codekeywords},
  identifierstyle=\color{black},
  stringstyle=\color{codestrings},
  commentstyle=\color{codecomment},
  columns=fullflexible,
  abovecaptionskip=\medskipamount,
  belowcaptionskip=\medskipamount,
  backgroundcolor=\color{codebackground},
}
\lstdefinestyle{clinestyle}{
tabsize=2,
frame=lines,
inputencoding=utf8,
language=C++,
keywordstyle=\bfseries\color{codekeywords},
identifierstyle=\color{black},
stringstyle=\color{codestrings},
commentstyle=\color{codecomment},
basicstyle=\ttfamily,
}
\lstnewenvironment{cblock}{\lstset{style=cblockstyle}}{}
\lstnewenvironment{clinee}{\lstset{style=clinestyle}}{}
\lstnewenvironment{pblock}{\lstset{style=pblockstyle}}{}

\makeatother

\usepackage{babel}
\begin{document}

\title{ViZDoom Competitions: Playing Doom from Pixels}

\author{\IEEEauthorblockN{Marek~Wydmuch\IEEEauthorrefmark{1}, Micha\l ~Kempka\IEEEauthorrefmark{1}
\& Wojciech~Ja\'{s}kowski\IEEEauthorrefmark{2}\IEEEauthorrefmark{1}}\IEEEauthorblockA{\IEEEauthorrefmark{1}Institute
of Computing Science, Poznan University of Technology, Pozna\'{n},
Poland\\
 \IEEEauthorrefmark{2}NNAISENSE SA, Lugano, Switzerland\\
mwydmuch@cs.put.poznan.pl, mkempka@cs.put.poznan.pl, wojciech@nnaisense.com}}
\maketitle
\begin{abstract}
This paper presents the first two editions of Visual Doom AI Competition
, held in 2016 and 2017. The challenge was to create bots that compete
in a multi-player deathmatch in a first-person shooter (FPS) game,
Doom. The bots had to make their decisions based solely on visual
information, i.e., a raw screen buffer. To play well, the bots needed
to understand their surroundings, navigate, explore, and handle the
opponents at the same time. These aspects, together with the competitive
multi-agent aspect of the game, make the competition a unique platform
for evaluating the state of the art reinforcement learning algorithms.
The paper discusses the rules, solutions, results, and statistics
that give insight into the agents' behaviors. Best-performing agents
are described in more detail. The results of the competition lead
to the conclusion that, although reinforcement learning can produce
capable Doom bots, they still are not yet able to successfully compete
against humans in this game. The paper also revisits the ViZDoom environment,
which is a flexible, easy to use, and efficient 3D platform for research
for vision-based reinforcement learning, based on a well-recognized
first-person perspective game Doom.

\end{abstract}


\section{Introduction\label{sec:Introduction}}

Since the beginning of the development of AI systems, games have been
natural benchmarks for AI algorithms because they provide well-defined
rules and allow for easy evaluation of the agent's performance. The
number of games solved by AI algorithms has increased rapidly in recent
years, and algorithms like AlphaGo~\cite{SilverHuangEtAl16nature,DBLP:journals/corr/abs-1712-01815}
beat the best human players in more and more complex board games that
have been previously deemed too sophisticated for computers. We have
also witnessed major successes in applying Deep Reinforcement Learning
to play arcade games~\cite{mnih2015humanlevel,DBLP:journals/corr/MnihBMGLHSK16,DBLP:journals/corr/PritzelUSBVHWB17},
for some of which machines, yet again, surpass humans. However, AI
agents faced with complex first-person-perspective, 3D environments
do not yet come even close to human performance. The disparity is
most striking when simultaneous use of multiple skills is required,
e.g., navigation, localization, memory, self-awareness, exploration,
or precision. Obtaining these skills is particularly important considering
the potential applicability of self-learning systems for robots acting
in the real world. 

Despite the limited real-world applicability still, a large body of
the research in reinforcement learning has concentrated on 2D Atari-like
games and abstract classical games. This is caused, in part, by the
scarcity of suitable environments and established benchmarks for harder
environments.

Introduced in early 2016, Doom-based ViZDoom~\cite{Kempka2016ViZDoom}
was the first published environment that aimed to provide a complex
3D first-person perspective platform for Reinforcement Learning (RL)
research. ViZDoom was created as a response to the sparsity of studies
on RL in complex 3D environments from raw visual information. The
flexibility of the platform has led to dozens of research works in
RL that used it as an experimental platform. It has also triggered
the development of other realistic 3D worlds and platforms suitable
for machine learning research, which have appeared since the initial
release of ViZDoom, such as Quake-based DeepMind's Lab~\cite{DBLP:journals/corr/BeattieLTWWKLGV16}
and Minecraft-based Project Malmo~\cite{Johnson:2016:MPA:3061053.3061259},
which follow similar principles as ViZDoom. Related are also environments
that focus on the task of house navigation using raw visual information
with very realistic renderers: House3D~\cite{2018arXiv180102209W},
AI2-THOR~\cite{ai2thor}, HoME~\cite{DBLP:journals/corr/abs-1711-11017},
CHALET~\cite{2018arXiv180107357Y}, UnrealCV~\cite{qiu2017unrealcv},
however, most of this environments focus only on that particular task
and lack extensibility and flexibility.

An effective way to promote research in such complex environments
as ViZDoom is to organize open competitions. In this paper, we describe
the first two editions of the Visual Doom AI Competition (VDAIC) that
were held during the Conference on Computational Intelligence and
Games 2016 and 2017. The unique feature of this annual competition
is that the submitted bots compete in multi-player matches having
the screen buffer as the only source of information about the environment.
This task requires effective exploration and navigation through a
3D environment, gathering resources, dodging missiles and bullets,
and, last but not least, accurate shooting. Such setting also implies
sparse and often delayed rewards, which is a challenging case for
most of the popular RL methods. The competition seems to be one of
a kind, for the time, as it combines 3D vision with the multi-player
aspect. The competition can be termed as belonging to `AI e-sport',
trying to merge the trending e-sports and events such as the driverless
car racing.

\section{The ViZDoom Research Platform\label{sec:Platform}}

\subsection{Design Requirements}

\begin{figure}
\centering{}\includegraphics[width=0.8\columnwidth]{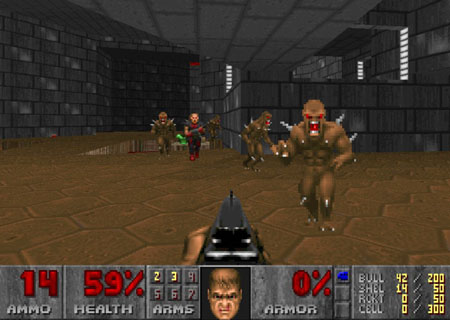}
\caption{\label{fig:doom}A sample screen from Doom showing the first-person
perspective.}
\end{figure}

The ViZDoom reinforcement learning research platform has been developed
to fulfill the following requirements:
\begin{enumerate}
\item based on popular open-source 3D FPS game (ability to modify and publish
the code),
\item lightweight (portability and the ability to run multiple instances
on a single machine with minimal computational resources),
\item fast (the game engine should not be the learning bottleneck by being
capable of generating samples hundreds or thousands of times faster
than real-time),
\item total control over the game's processing (so that the game can wait
for the bot decisions or the agent can learn by observing a human
playing),
\item customizable resolution and rendering parameters,
\item multi-player games capabilities (agent vs. agent, agent vs. human
and cooperation),
\item easy-to-use tools to create custom scenarios, 
\item scripting language to be able to create diverse tasks,
\item ability to bind different programming languages (preferably written
in C++),
\item multi-platform.
\end{enumerate}
In order to meet the above-listed criteria, we have analyzed several
recognizable FPS game engines: Doom, Doom 3, Quake III, Half-Life
2, Unreal Tournament and Cube. Doom (see Fig.~\ref{fig:doom}) with
its low system requirements, simple architecture, multi-platform and
single and multi-player modes met most of the conditions (see~\cite{Kempka2016ViZDoom}
for a detailed analysis) and allowed to implement features that would
be barely achievable in other game engines, e.g., off-screen rendering,
efficiency, and easy-to-create custom scenarios. The game is highly
recognizable and runs on the three major operating systems. It was
also designed to work in $320\times240$ resolution and despite the
fact that modern implementations allow higher resolutions, it still
utilizes low-resolution textures, which positively impacts its resource
requirements.

The nowadays unique feature of Doom is its software renderer. This
is especially important for reinforcement learning algorithms, which
are distributed on CPUs rather than on GPUs. Yet another advantage
of the CPU rendering is that Doom can effortlessly be run without
a desktop environment (e.g., remotely, in a terminal) and accessing
the screen buffer does not require transferring it from the graphics
card.

Technically, ViZDoom is based on the modernized, open-source version
of Doom's original engine, ZDoom\footnote{\url{zdoom.org}}, which
has been actively supported and developed since 1998. The large community
gathered around the game and the engine has provided a lot of tools
that facilitate creating custom scenarios.

\subsection{Features}

ViZDoom provides features that can be exploited in a wide range of
AI and, in particular, machine learning experiments. It allows for
different control modes, custom scenarios and access to additional
information concerning the scene, including per-pixel depth (depth
buffer), visible objects, and a top-down view map. In the following
sections, we list the most important features of ViZDoom 1.1.5, which
substantially extend the features of the initial 1.0 version~\cite{Kempka2016ViZDoom}.

\subsubsection{Control Modes}

ViZDoom provides four control modes: i) \emph{synchronous} \emph{player},
ii) \emph{synchronous} \emph{spectator}, iii) \emph{asynchronous}
\emph{player}, and iv) \emph{asynchronous} \emph{spectator}.

In \emph{asynchronous} modes, the game runs at constant $35$ frames
per second and if the agent reacts too slowly, it can miss one or
more frames. Conversely, if it makes a decision too quickly, it is
blocked until the next frame arrives from the engine. Thus, for the
purpose of reinforcement learning research, it is more efficient to
use \emph{synchronous} modes, in which the game engine waits for the
decision maker. This way, the learning system can learn at its pace,
and it is not limited by any temporal constraints.

Importantly, for experimental reproducibility and debugging purposes,
the synchronous modes run fully deterministically.

In the player modes, it is the agent who makes actions during the
game. In contrast, in the \emph{spectator} modes, a human player is
in control, and the agent only observes the player's actions.

ViZDoom provides support both for \emph{single-} and \emph{multi-player}
games, which accept up to sixteen agents playing simultaneously on
the same map communicating over a network. As of the 1.1 version of
ViZDoom, multi-player games can be run in all modes (including the
synchronous ones). Multi-player can involve deathmatch, team deathmatch,
or fully cooperative scenarios.

\subsubsection{Scenarios}

One of the most important features of ViZDoom is the ability to execute
custom scenarios, which are not limited to just playing Doom. This
includes creating appropriate maps (``what the world looks like''),
programming the environment's mechanics (``when and how things happen''),
defining terminal conditions (e.g., ``killing a certain monster'',
``getting to a certain place'', ``getting killed''), and rewards
(e.g., for ``killing a monster'', ``getting hurt'', ``picking
up an object''). This mechanism opens endless experimentation possibilities.
In particular, it allows researchers to create scenarios of difficulty
on par with  capabilities of the state-of-the-art learning algorithms.

Creation of scenarios and maps is possible thanks to easy-to-use software
tools developed by the Doom community. The two recommended free tools
are Doom~Builder~2\footnote{\url{http://www.doombuilder.com}} and
SLADE~3\footnote{\url{http://slade.mancubus.net}}. Both visual editors
make it easy to define custom maps and coding the game mechanics in
Action Code Script for both single- and multi-player games. They also
enable to conveniently test a scenario without leaving the editor.

While any rewards and constraints can be implemented by using the
scripting language, ViZDoom provides a direct way for setting most
typical kinds of rewards (e.g., for \textquotedblleft living\textquotedblright{}
or ``dying''), constraints regarding the elementary actions/keys
that can be used by agent, or temporal constraints such as the maximum
episode duration. Scenarios do not affect any rendering options (e.g.,
screen resolution, or the crosshair visibility), which can be customized
in configuration files or via the API.

ViZDoom comes with more than a dozen predefined scenarios allowing
for the training of fundamental skills like shooting or navigation.

\subsubsection{Automatic Objects Labeling}

The objects in the current view can be automatically labeled and provided
as a separate input channel together with additional information about
them (cf.~Fig.~\ref{fig:zbuffer}). The environment also provides
access to a list of labeled objects (bounding boxes, their names,
positions, orientations, and directions of movement). This feature
can be used for supervised learning research.

\subsubsection{Depth Buffer}

\begin{figure*}
\centering{}\includegraphics[width=1\linewidth]{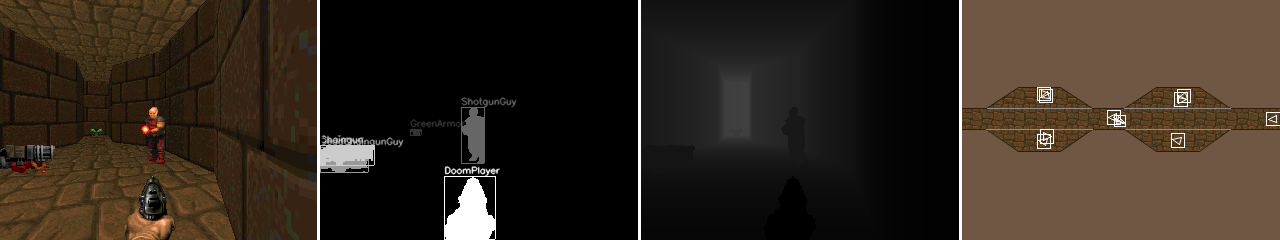}\caption{\label{fig:zbuffer}Apart from the regular screen buffer, ViZDoom
provides access to a buffer with labeled objects, a depth buffer,
and a top-down view map. Note that for competitions, during the evaluations,
the agents were provided with only the standard (left-most) view.}
\end{figure*}

ViZDoom provides access to the renderer's depth buffer (see Fig.~\ref{fig:zbuffer}),
which may be used to simulate the distance sensors common in mobile
robots and help an agent to understand the received visual information.
This feature gives an opportunity to test whether an agent can autonomously
learn the whereabouts of the objects in the environment.

\subsubsection{Top-Down View Map}

ViZDoom can also render a top-down representation (a map) of the episode's
environment. The map can be set up to display the entire environment
or only the part already discovered by the agent. In addition, it
can show objects including their facings and sizes. This feature can
be used to test whether the agents can efficiently navigate in a complex
3D space, which is a common scenario in mobile robotics research.
Optionally, it allows turning ViZDoom into a 2D environment, which
eliminates the need for an auxiliary 2D environment for simpler scenarios.

\subsubsection{Off-Screen Rendering and Frame Skipping\label{subsec:Off-Screen-Skipping}}

To facilitate computationally heavy machine learning experiments,
ViZDoom is equipped with an off-screen rendering and frame skipping
features. The off-screen rendering decreases the performance burden
of displaying the game on the screen and makes it possible to run
experiments on the servers (no graphical interface required). Frame
skipping, on the other hand, allows omitting to render some frames
completely since typically an effective bot does not need to see every
single frame.

\subsubsection{ViZDoom's Performance}

Our performance tests show that ViZDoom can render up to $2500$ frames
per second on average in most of the scenarios in $320\times240$
and even $7000$ frames per second in the low-resolution of $160\times120$ on
a modern CPU (single threaded).  

The main factors affecting the ViZDoom performance are the rendering
resolution, and computing the additional buffers (depth, labels, and
the top-down view map). In the case of low resolutions, the time needed
to render one frame is negligible compared to the backpropagation
time of any reasonably complex neural network. It is also worth mentioning
that one instance of ViZDoom requires only a dozen MBs of RAM, which
allows running many instances simultaneously.

\subsubsection{Recording and replaying episodes}

Last but not least, the ViZDoom games can be effortlessly recorded
and saved to disk to be later replayed. During playback, all buffers,
rewards, game variables, and the player's actions can be accessed
just like in the spectator mode, which becomes useful for learning
from demonstration. Moreover, the rendering settings (e.g., the resolution,
textures, etc.) can be changed at the replay time. This is useful
for preparing high-resolution demonstration movies.

\subsection{Application Programming Interface (API)\label{subsec:api}}

ViZDoom API is flexible and easy-to-use. It was designed to conveniently
support reinforcement learning and learning from demonstration experiments,
and, therefore, it provides full control over the underlying Doom
process. In particular, it is possible to retrieve the game's screen
buffer and make actions that correspond to keyboard buttons (or their
combinations) and mouse actions. Some of game state variables such
as the player's health or ammunition are available directly.

The ViZDoom's API was written in C++. The API offers a myriad of configuration
options such as control modes and rendering options. In addition to
the C++ support, bindings for Python, Lua, Java and Julia have been
provided. A sample code using the Python API with a randomly behaving
bot is shown in Fig.~\ref{fig:PythonExample}.

\begin{figure}
\begin{pblock}
import vizdoom as vzd
from random import choice
game = vzd.DoomGame()
game.load_config("custom_config.cfg")
game.add_game_args("+name RandomBot")
game.init()
# Three sample actions: turn left/right and shoot
actions = [[1, 0, 0], [0, 1, 0], [0, 0, 1]]
while not game.is_episode_finished():
    if game.is_player_dead():
        game.respawn_player()    
    # Retrieve the state
    state = game.get_state()
    screen = state.screen_buffer
    health, ammo = state.game_variables
    game.make_action(choice(actions))
\end{pblock}\caption{\label{fig:PythonExample}A sample random agent in Python.}
\end{figure}

\section{Visual RL Research Platforms\label{sec:env-compare}}

The growing interest in Machine Learning and Reinforcement Learning
that had given rise to ViZDoom has recently triggered development
of a number of other environments designed for RL experimentations. 

DeepMind Lab~\cite{DBLP:journals/corr/BeattieLTWWKLGV16} and Project
Malmo~\cite{Johnson:2016:MPA:3061053.3061259} are the closest counterparts
of ViZDoom since they both involve the first-person perspective. Similarly
to ViZDoom, they are based on popular games. Project Malmo has been
developed on top of Minecraft while DeepMind Lab uses Quake 3 Arena,
which was also considered for the ViZDoom project. It has been, however,
rejected due to limited scripting capabilities and a developer unfriendly
server-client architecture, which also limits the performance of the
environment. DeepMind Lab vastly extends scripting capabilities of
Quake and adds custom resources for the project that overhauls the
look of the environment, which makes DeepMind Lab more detached from
its base game compared to ViZDoom.

All three platforms allow for defining custom scenarios. Project Malmo,
is particularly flexible in this respect, giving a lot of freedom
by providing a full access to the state of the environment during
runtime. Unfortunately, in contrast to ViZDoom, there is no visual
editor available for Project Malmo. DeepMind Lab's scripting allows
for building complex environments on top of it (e.g., Psychlab~\cite{DBLP:journals/corr/abs-1801-08116}).
For creating custom geometry, original Quake 3 Arena visual editor
can be adopted but it is burdensome in use. On the other hand, ViZDoom
is compatible with accessible Doom's community tools that allow for
creating levels and scripting them. However, due to the engine limitations,
creating some types of scenarios may require more work compared to
the two previous environments (e.g., scenarios with a lot of randomized
level geometry).

All three platforms provide an RL friendly API in a few programming
languages and provide depth buffer access but only   ViZDoom makes
it possible to obtain the detailed information about objects visible
on the screen, the depth buffer, and offers a top-down view map.

UnrealCV~\cite{qiu2016unrealcv} is yet another interesting project
that offers a high-quality 3D rendering. It provides an API for Unreal
Engine 4, that enables users to obtain the environment state, including
not only the rendered image but also a whole set of different auxiliary
scene information such as the depth buffer or the scene objects. UnrealCV
is not a self-contained environment with existing game mechanics and
resources, it must be attached to an Unreal Engine-based game. This
characteristic allows creating custom scenarios as separate games
directly in Unreal Engine Editor using its robust visual scripting
tools. However, since UnrealCV is designed to be a universal tool
for computer vision research, it does not provide any RL-specific
abstraction layer in its API. 

Unity ML-Agents\footnote{\url{https://unity3d.com/machine-learning}}
is the most recent project, which is still in development (currently
in beta). It follows a similar principle as UnrealCV providing a Python
API for scenarios that can be created using the Unity engine. However,
like Project Malmo, it aims to be a more general RL platform with
a flexible API. It allows a user to create a wide range of scenarios,
including learning from a visual information.

While the ViZDoom graphic is the most simplistic among all major first-person
perspective environments, it also makes it very lightweight, allowing
to run multiple instances of the environment using only the small
amount of available computational resources. Among the available environments,
it is the most computationally efficient, which is an important practical
experimentation aspect. A detailed comparison of environments can
be found in Table~\ref{tab:environments}.

\begin{table*}
\begin{centering}
\caption{\label{tab:environments}Overview of 3D First-Person Perspective RL
platforms.}
\par\end{centering}
\setlength\tabcolsep{4pt}
\resizebox{\textwidth}{!}{%
\begin{centering}
\begin{tabular}{ccccccc}
\toprule 
\multirow{1}{*}{Feature} & ViZDoom & DeepMind Lab & Project Malmo & OpenAI Universe & UnrealCV & Unity ML-Agents\tabularnewline
\midrule
Base Game/Engine & Doom/ZDoom & Quake 3/ioquake3 & Minecraft & Many & Unreal Engine 4 & Unity\tabularnewline
Public release date & March 2016 & December 2016 & May 2016 & December 2016 & October 2016 & September 2017\tabularnewline
Open-source & \OK & \OK & \OK & $\dagger$ & \OK & $\ddagger$\tabularnewline
Language & C++ & C & Java & Python & C++ & C\#, Python\tabularnewline
API languages & Python, Lua, C++, Java, Julia & Python, Lua & Python, Lua, C++, C\#, Java & Python & Python, MatLab & Python\tabularnewline
\midrule
Windows & \OK &  & \OK & \OK & \OK & \OK\tabularnewline
Linux & \OK & \OK & \OK & \OK & \OK & \OK\tabularnewline
Mac OS & \OK & \OK & \OK & \OK & \OK & \OK\tabularnewline
\midrule
Game customization capabilities & \OK & \OK & Limited & - & \OK & \OK\tabularnewline
Scenario editing tools & Visual Editors & Text + Visual Editor & XML defined & - & Unreal Engine 4 Editor & Unity Editor\tabularnewline
Scenario scripting language & Action Code Script & Lua & Controlled via API & - & Unreal Engine 4 Blueprints & C\#, JavaScript\tabularnewline
\midrule
Framerate at $320\times240$ & 2500 (CPU) & 100 (CPU)/800 (GPU) & 120 (GPU) & 60 (GPU, locked) & Depends & Depends\tabularnewline
Depth buffer & \OK & \OK & \OK &  & \OK & \tabularnewline
Auto object labeling & \OK &  &  &  & \OK & \tabularnewline
Top-down map & \OK &  &  &  &  & \tabularnewline
\midrule
Low level API & \OK & \OK & \OK &  & \OK & \tabularnewline
RL friendly abstraction layer & \OK & \OK & \OK & \OK &  & \OK\tabularnewline
Multi-player support & \OK & \OK & \OK &  &  & \tabularnewline
\midrule 
System requirements & Low & Medium & Medium & High$\ast$ & High & Medium$\ast$\tabularnewline
\bottomrule
\end{tabular}
\par\end{centering}
}\vspace{4pt}

$\ast$ -- platform allows creating scenarios with varying graphical
level and thus requirements, $\dagger$ -- platform is open-source,
however code of the most base games are closed, $\ddagger$ -- platfrom
is open-source, however Unity engine code is closed.
\end{table*}

\section{Visual Doom AI Competitions (VDAIC) \label{sec:Competitions}}

\subsection{Motivation}

Doom has been considered one of the most influential titles in the
game industry since it had popularized the first-person shooter (FPS)
genre and pioneered immersive 3D graphics. Even though more than 25
years have passed since Doom\textquoteright s release, the methods
for developing AI bots have not improved \foreignlanguage{english}{qualitatively}
in newer FPS productions. In particular, bots still need to \textquotedblleft cheat\textquotedblright{}
by accessing the game\textquoteright s internal data such as maps,
locations of objects and positions of (player or non-player) characters
and various metadata. In contrast, a human can play FPS games using
a computer screen as the sole source of information, although the
sound effects might also be helpful. In order to encourage development
of bots that act only on raw visual information and to evaluate the
state of the art of visual reinforcement learning, two AI bot competitions
have been organized at the IEEE Conference on Computational Intelligence
and Games 2016 and 2017.

\subsection{Other AI competitions}

There have been many game-based AI contests in the past~\cite{togelius2016run}.
The recent AI competition examples include General Video Game (GVGAI)~\cite{b681c201aaa3488aa06c82bb01ef2787},
Starcraft~\cite{AIIDE1715830}, Pac-Mac~\cite{7860446},  and the
Text-Based Adventure~\cite{2018arXiv180801262A}. Each of the competitions
provides a different set of features and constraints.

Most of the contests give access to high-level representations of
game states, which are usually discrete. VDAIC is uncommon here since
it requires playing only and directly from raw high-dimensional pixel
information representing a 3D scene (screen buffer).

Many competitions concentrate on planning. For instance, GVGAI provides
an API that allows to sample from a forward model of a game. This
turned the competition into an excellent benchmark for variants of
Monte Carlo Tree Search. 

The Starcraft AI competition shares the real-time aspect of VDAIC
but, similarly to GVGAI, it focuses on planning basing on high-level
state representations. This is reflected in the proposed solutions~\cite{AIIDE1715830},
which involve state search and hand-crafted strategies. Few competitions
like Learning to Run~\cite{DBLP:journals/corr/abs-1804-00361} and
the learning track of GVGAI target model-free environments (typical
RL settings). However,  both of them  still provide access to relatively
high-level observations.

It is apparent that the Visual Doom AI Competition has been unique
and has filled a gap in the landscape of AI challenges by requiring
bots to both perceive and plan in real-time in a 3D multi-agent environment. 

\subsection{Edition 2016}

\subsubsection{Rules }

The task of the competition was to develop a bot that competes in
a multi-player deathmatch game with the aim to maximize the number
of frags, which, by Doom's definition, is the number of killed opponents
decreased by the number of committed suicides (bot dies due to a damages
inflicted by its own actions). The participants of the competition
were allowed to prepare their bots offline and use any training method,
external resources like custom maps, and all features of the environment
(e.g., depth buffer, custom game variables, Doom's built-in bots).
However, during the contest, bots were allowed to use only the screen
buffer (the left-most view in~Fig.~\ref{fig:zbuffer}) and information
available on the HUD such as ammunition supply, health points left,
etc. Participants could configure the screen format (resolution, colors)
and rendering options (crosshair, blood, HUD visibility, etc.). All
features of the environment providing the bots information typically
inaccessible to human players were blocked. The participants were
allowed to choose between two sets of textures: the original ones
and freeware substitutes. 

For evaluation, the asynchronous multi-player mode was used (a real-time
game with 35 frames per second). Each bot had a single computer at
its exclusive disposal (Intel(R) Core(TM) i7-4790 CPU @ 3.60GHz, 16GB
RAM with Nvidia GTX 960 4GB). Participants could choose either Ubuntu
16.04 or Windows 10 and provide their code in Python, C++, or Java.

The competition consisted of two independent tracks (see Section~\ref{sec:tracks}).
Each of them consisted of 12 matches lasting 10 minutes (2 hours of
gameplay per track in total). Bots started every match and were respawned
immediately after death at one of the respawn points, selected as
far as possible from the other players. Additionally, bots were invulnerable
to attacks for the first two seconds after a respawning.

\subsubsection{Tracks\label{sec:tracks}}

\paragraph{Track 1: Limited Deathmatch on a Known Map}

The agents competed on a single map, known in advance. The only available
weapon was a rocket launcher, with which the agents were initially
equipped. The map consisted mostly of relatively constricted spaces,
which allow killing an opponent by hitting a nearby wall with a rocket
(blast damage). For the same reason, it was relatively easy to kill
oneself. The map contained resources such as ammunition, medikits,
and armors. Due to the fact that the number of participants (9) exceeded
the ViZDoom's 1.0 upper limit of 8 players per game, a fair matchmaking
scheme was developed. For each of the first 9 matches, a single bot
was excluded. For the remaining 3 matches, 2 bots that had performed
worst in the first 9 matches were excluded.

\paragraph{Track 2: Full Deathmatch on Unknown Maps}

\begin{figure*}
\centering{}\includegraphics[width=1\linewidth]{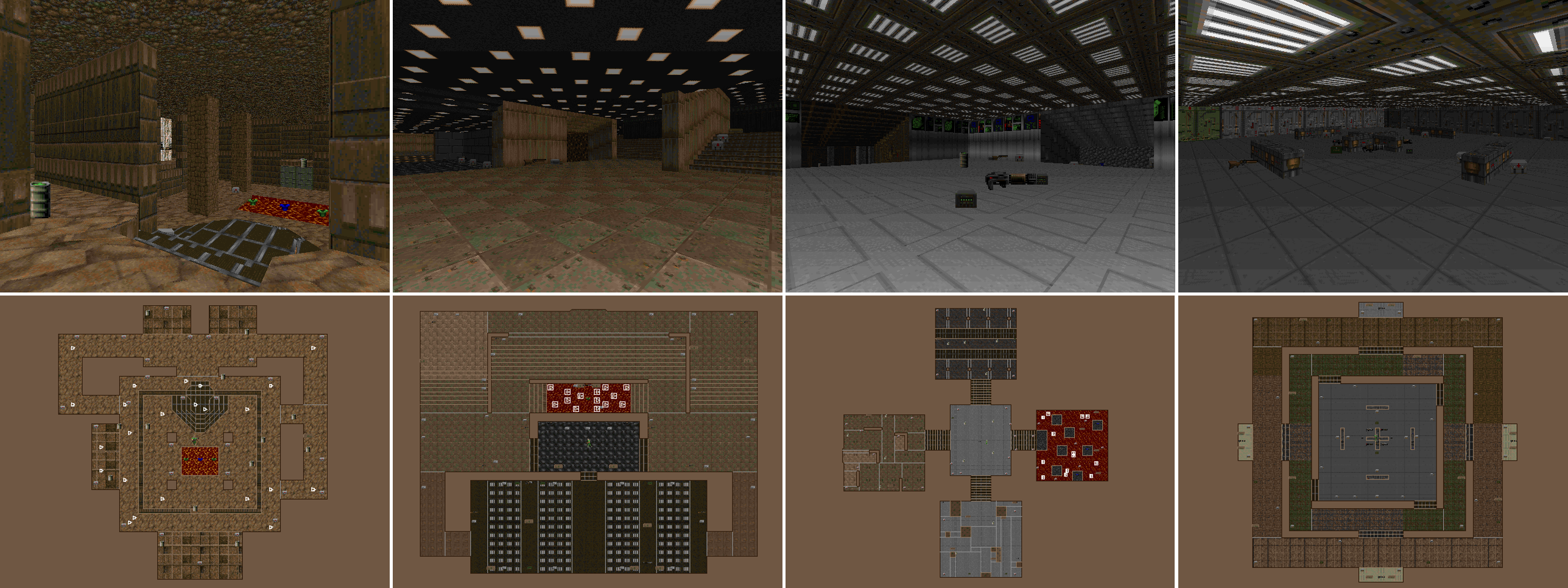}
\caption{\label{fig:cig_2016_maps}The map used for evaluation in the Track
1 (left), and three maps used in Track 2 in the 2016 edition of Visual
Doom AI Competition.}
\end{figure*}

The agents competed four times on each of the three previously unseen
maps (see Fig.~\ref{fig:cig_2016_maps}), and were initially equipped
with pistols. The maps were relatively spacious and contained open
spaces, which made accurate aiming more relevant than in Track 1.
The maps contained various weapons and items such as ammunition, medikits,
and armors. A sample map was provided. All maps were prepared by authors
of the competition.

Notice that Track 2 has been considerably harder than Track~1. During
the evaluation, agents were faced with completely new maps, so they
could not learn the environment by heart during the training as in
Track 1. And while it is enough to move randomly and aim well to be
fairly effective for Track 1, a competent player for Track 2 should
make strategic decisions such as where to go, which weapon to use,
explore or wait, etc.

\subsubsection{Results}

The results of the competition are shown in Tables~\ref{tab:results-track1-2016}~and~\ref{tab:results-track2-2016},
for Track 1 and 2, respectively. For future reference, all matches
were recorded and are available publicly (see Appendix).

\paragraph{Track 1}

The bots in 7 out of 9 submissions were competent enough to systematically
eliminate the opponents. Among them, four bots stand out by scoring
more than 300 frags: F1, Arnold, Clyde and TUHO. The difference between
the second (Arnold) and the third (Clyde) place was minuscule (413
vs. 393 frags) and it is questionable whether the order remained the
same if games were repeated. There is no doubt, however, that F1 was
the best bot beating the forerunner (Arnold) by a large margin. F1
was also characterized by the least number of suicides. Note, however,
that generally, the number of suicides is high for all the agents.
Interestingly, despite the fact that F1 scored the best, it was Arnold
who was gunned down the least often.

\begin{table}
\begin{centering}
\caption{\label{tab:results-track1-2016}Results of the 2016 Competition: Track
1. `Frags' is the number of opponent kills decreased by the number
of suicides of the agent. `F/D' denotes Frags/Death. `Deaths' include
suicides.}
\par\end{centering}
\setlength\tabcolsep{4pt}
\centering{}%
\begin{tabular*}{1\linewidth}{@{\extracolsep{\fill}}ccrrrrr}
\toprule 
Place & Bot & Frags & F/D ratio & Kills & Suicides & Deaths\tabularnewline
\midrule
1 & F1 & \textbf{559} & 1.35 & \textbf{597} & \textbf{38} & 413\tabularnewline
2 & Arnold & 413 & \textbf{1.90} & 532 & 119 & \textbf{217}\tabularnewline
3 & Clyde & 393 & 0.77 & 476 & 83 & 509\tabularnewline
4 & TUHO & 312 & 0.67 & 424 & 112 & 465\tabularnewline
5 & 5vision & 142 & 0.28 & 206 & 64 & 497\tabularnewline
6 & ColbyMules & 131 & 0.25 & 222 & 91 & 516\tabularnewline
7 & AbyssII & 118 & 0.21 & 217 & 99 & 542\tabularnewline
8 & WallDestroyerXxx & -130 & -0.41 & \emph{13} & 143 & 315\tabularnewline
9 & Ivomi & \emph{-578} & \emph{-0.68} & 149 & \emph{727} & \emph{838}\tabularnewline
\bottomrule
\end{tabular*}
\end{table}

\paragraph{Track 2}

In Track 2, IntelAct was the best bot, significantly surpassing its
competitors on all maps. Arnold, who finished in the second place,
was killed the least frequently. Compared to Track 1, the numbers
of kills and suicides (see Tables \ref{tab:results-track1-2016} and
\ref{tab:results-track2-2016}) are significantly lower, which is
due to less usage of rocket launchers that are dangerous not only
for the opponent but also for the owner.

\begin{table*}
\begin{centering}
\caption{\label{tab:results-track2-2016}Results of the 2016 Competition: Track
2. `M' denotes map and `T' denotes a total statistic.}
\par\end{centering}
\centering{}\setlength\tabcolsep{6pt}%
\begin{tabular*}{1\linewidth}{@{\extracolsep{\fill}}ccccccccccccccccccc}
\toprule 
\multirow{2}{*}{Place} & \multirow{2}{*}{Bot} & \multirow{2}{*}{Total Frags} & \multirow{2}{*}{F/D ratio} &  & \multicolumn{4}{c}{Kills} &  & \multicolumn{4}{c}{Suicides} &  & \multicolumn{4}{c}{Deaths}\tabularnewline
\cmidrule{6-19} 
 &  &  &  &  & M1 & M2 & M3 & T &  & M1 & M2 & M3 & T &  & M1 & M2 & M3 & T\tabularnewline
\midrule
1 & IntelAct & \textbf{256} & 3.08 &  & \textbf{113} & \textbf{49} & \textbf{135} & \textbf{297} &  & \emph{19} & \emph{17} & 5 & \emph{41} &  & 47 & 24 & 12 & 83\tabularnewline
2 & Arnold & 164 & \textbf{32.8} &  & 76 & 37 & 53 & 167 &  & 2 & \textbf{1} & \textbf{0} & \textbf{3} &  & \textbf{3} & \textbf{1} & \textbf{1} & \textbf{5}\tabularnewline
3 & TUHO & 51 & 0.66 &  & 51 & 9 & 13 & 73 &  & 7 & 15 & \textbf{0} & 22 &  & 31 & 29 & 17 & 77\tabularnewline
4 & ColbyMules & 18 & 0.13 &  & 8 & 5 & 13 & 26 &  & 1 & 7 & \textbf{0} & 8 &  & 60 & 27 & 44 & 129\tabularnewline
5 & 5vision & 12 & 0.09 &  & 12 & 10 & 4 & 26 &  & 3 & 8 & 3 & 14 &  & 45 & \emph{37} & 47 & 131\tabularnewline
6 & Ivomi & -2 & -0.01 &  & 6 & 5 & 2 & 13 &  & 2 & 13 & \textbf{0} & 15 &  & \emph{69} & 33 & 35 & 137\tabularnewline
7 & WallDestroyerXxx & \emph{-9} & \emph{-0.06} &  & \emph{2} & \emph{0} & \emph{0} & \emph{2} &  & \textbf{0} & 5 & \emph{6} & 11 &  & 48 & 30 & \emph{78} & \emph{156}\tabularnewline
\bottomrule
\end{tabular*}
\end{table*}

\subsubsection{Notable Solutions}

\begin{table}
\caption{\label{tab:frameworks_2016}Algorithms and Frameworks Used in the
2016 Competition}

\centering{}%
\begin{tabular*}{1\linewidth}{@{\extracolsep{\fill}}>{\centering}p{0.2\linewidth}>{\centering}p{0.3\linewidth}>{\centering}p{0.4\linewidth}}
\toprule 
Bot & Framework used & Algorithm\tabularnewline
\midrule
IntelAct & Tensorflow & Direct Future Prediction (DFP)\tabularnewline
F1 & Tensorflow + Tensorpack & A3C, curriculum learning\tabularnewline
Arnold & Theano  & DQN, DRQN\tabularnewline
Clyde & Tensorflow & A3C\tabularnewline
TUHO & Theano + Lasagne & DQN + Haar Detector\tabularnewline
5vision & Theano + Lasagne & DARQN \cite{DBLP:journals/corr/SorokinSPFI15}\tabularnewline
ColbyMules & Neon & unknown\tabularnewline
AbyssII & Tensorflow & A3C\tabularnewline
Ivomi & Theano + Lasagne & DQN\tabularnewline
WallDestroyerXxx & Chainer & unknown\tabularnewline
\bottomrule
\end{tabular*}
\end{table}

Table~\ref{tab:frameworks_2016} contains a list of bots submitted
to the competition. Below, the training methods for the main bots
are briefly described:
\begin{itemize}
\item \textbf{F1} (Yuxin Wu, Yuandong Tian) - the winning bot of Track 1
was trained with a variant of the A3C algorithm~\cite{DBLP:journals/corr/MnihBMGLHSK16}
with curriculum learning~\cite{Bengio:2009:CL:1553374.1553380}.
The agent was first trained on an easier task (weak opponents, smaller
map) to gradually face harder problems (stronger opponents, bigger
maps). Additionally, some behaviors were hardcoded (e.g., increased
movement speed when not firing)~\cite{Wu2017TrainingAF}. 
\item \textbf{IntelAct} (Alexey Dosovitskiy, Vladlen Koltun) - the best
agent in Track 2 was trained with Direct Future Prediction (DFP, \cite{DBLP:journals/corr/DosovitskiyK16}).
The algorithm is similar to DQN \cite{mnih2015humanlevel} but instead
of maximizing the expected reward, the agent tries to predict future
measurement vector (e.g., health, ammunition) for each action based
on the current measurement vector and the environment state. The agent's
actual goal is defined as a linear combination of the future measurements
and can be changed on the fly without retraining. Except for weapon
selection, which was hardcoded, all behaviors were learned from playing
against bots on a number of different maps. The core idea of DFP is
related to UNREAL \cite{DBLP:journals/corr/JaderbergMCSLSK16} as,
in addition to the reward, it predicts auxiliary signals.
\item \textbf{Arnold} (Guillaume Lample, Devendra Singh Chaplot) took the
second place in both tracks (two sets of parameters and the same code).
The training algorithm~\cite{DBLP:journals/corr/LampleC16} contained
two modules: navigation, obtained with DQN, and aiming, trained with
Deep Recurrent Q-Learning (DRQN~\cite{DBLP:journals/corr/HausknechtS15}).
Additionally, the aiming network contains an additional output with
a binary classifier indicating whether there is an enemy visible on
the screen. Based on the classifier decision, either the navigation
or the aiming network decided upon the next action. The navigation
network was rewarded for speed and penalized for stepping on lava
whereas the aiming network was rewarded for killing, picking up objects,
and penalized for losing health or ammunition. It is worth noting
that Arnold crouched constantly, which gave him a significant advantage
over the opponents. This is reflected in his high frags to deaths
ratio especially for Track 2 (see Tables~\ref{tab:results-track1-2016}~and~\ref{tab:results-track2-2016}).
Arnold was trained on a set of maps created by the solution's authors,
starting with maps full of weak enemies. 
\item \textbf{Clyde} (Dino Ratcliffe), the bot that took the third place
in Track 1, was trained with the vanilla A3C algorithm rewarded for
killing, collecting items and penalized for suicides. For more details
cf.~\cite{AAAIW1715130}. 
\item \textbf{TUHO }(Anssi `Miffyli' Kanervisto, Ville Hautam{\"a}ki),
the bot that took the third place in Track 2 was trained similarly
to Arnold. TUHO used two independent modules. The navigation module
was trained with a Dueling DQN~\cite{DBLP:journals/corr/WangFL15},
rewarded for speed while the aiming network featured a classical Haar
detector \cite{Viola01rapidobject} trained on manually labeled examples.
If an enemy was detected in the frame, the agent turned appropriately
and fired. The navigation network was used otherwise. TUHO was trained
only on the two supplied maps and two other maps bundled with ViZDoom.
\end{itemize}

\subsubsection{Discussion}

Although the top bots were quite competent and easily coped with the
Doom's built-in bots, no agent came close to the human's competence
level. The bots were decent at tactical decisions (aiming and shooting)
but poor on a strategic level (defense, escaping, navigation, exploration).
This was especially visible for Track 2, where the maps were larger
and required a situational awareness -- bots often circled around
the same location, waiting for targets to appear. 

No agent has been capable of vertical aiming. That is why Arnold,
which hardcoded crouching, avoided many projectiles, achieving exceptionally
a high frags/death ratio. Other bots, probably, had never seen a crouching
player during their training and therefore were not able to react
appropriately. 

It was also observed that strafing (moving from side to side without
turning)  was not used very effectively (if at all) and agents did
not make any attempts to dodge missiles -- a feat performed easily
by humans. Bots did not seem to use any memory as well, as they did
not try to chase bots escaping from their field of view.

Most of the submitted agents were trained with the state-of-the-art
(as of 2016) RL algorithms such as A3C and DQN but the most successful
ones additionally addressed the problem of sparse, delayed rewards
by auxiliary signals (IntelAct, Arnold) or curriculum training (F1).

\subsubsection{Logistics}

Before the contest evaluation itself, three, optional warm-up rounds
were organized to accustom the participants to the submission and
evaluation process and give them a possibility to check their bots
effectiveness against each other. The performance of solutions was
tested on known maps and the results in the form of tabular data and
video recordings were made publicly available.

The participants were supposed to submit their code along with a list
of required dependencies to install. In terms of logistics, testing
the submissions was a quite a strenuous process for our small team
due to various compatibility issues and program dependencies especially
for solutions employing less standard frameworks. This caused an enormous
time overhead and created a need for more automated verification process.
That is why it was decided that Docker containers~\cite{Merkel:2014:DLL:2600239.2600241}
would be used for the subsequent edition of the competition, relieving
the organizers from dealing with the dependency and compatibility
issues.

\subsection{Edition 2017}

\subsubsection{Changes Compared to the Edition 2016}

\begin{figure*}
\centering{}\includegraphics[width=1\linewidth]{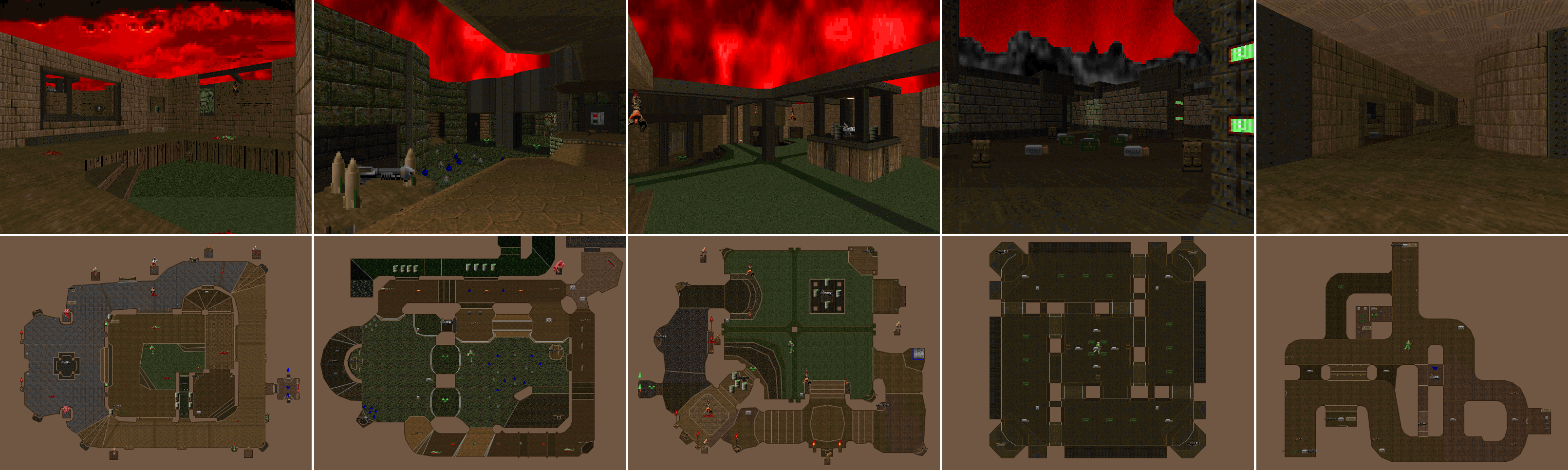}
\caption{\label{fig:cig_2017_maps}The five maps used for evaluation in Track
2 in 2017 competition.}
\end{figure*}

The rules and logistics of the 2017 competition did not differ much
from the ones of the previous edition. The changes were as follows:
\begin{itemize}
\item The new version of ViZDoom (1.1.2) was used as the competition environment;
all new features described in Section~\ref{sec:Platform} were allowed
for the training or preparation of the bots. The raised player limit
(from 8 to 16) made it possible to fit all the submitted bots in a
single game. 
\item Each track consisted of 10 matches, each lasting 10 minutes.
\item Track 2 consisted of 5 previously unseen maps (see Fig.~\ref{fig:cig_2017_maps}),
each one was used for 2 matches. Maps were chosen randomly from four
highly rated Doom multi-player map packs, each containing a number
of maps varying from 6 to 32 (which gives, in total, 88 maps), that
were selected from a more substantial list of map packs suggested
by the ZDoom community. Thus, the selected maps were characterized
by good quality and thoughtful design. They were also much smaller
compared to the maps used in the Track 2 of the 2016 competition,
leading to the more frequent interaction between the agents.
\item The respawn time (an obligatory waiting after death) of 10 seconds
was introduced to encourage gameplay that also focuses on the survival
instead of reckless killing and to limit the number of frags obtained
on weaker bots.
\item The crouching action was disabled as it gives an enormous advantage
over non-crouching players while being achievable by hardcoding of
a single key press (which was implemented in one of 2016's submissions).
A situation in which an agent learned to crouch effectively on its
own would arguably be an achievement but that was not the case.
\item Matches were initiated by a dedicated host program (for recording
purposes) and all agent's processes were run from Docker containers~\cite{Merkel:2014:DLL:2600239.2600241},
submitted by the participants.
\item The winning bots of the 2016 competition were added to each track
as baselines; they were also made available to the participants for
training or evaluation.
\item In the previous edition of the competition, most of the participants
did not join the warm-up rounds which made it even more difficult
for the organizers and also participants to estimate quantity and
quality of the final submissions. That is why in 2017, an obligatory
elimination round was introduced. Only teams that had participated
in the elimination rounds and presented sufficiently competent bots
were allowed to enter the final round.
\end{itemize}

\subsubsection{Results}

The results of the competition were shown in Tables~\ref{tab:results-track1-2017}
and~\ref{tab:results-track2-2017} for Track 1 and 2, respectively.
For this edition of the competition, the engine was extended to extract
additional information about agents performance, specifically: the
average movement speed (given in km/h, assuming that 128 of game units
correspond to 3 meters in the real world), the number of performed
attacks, their average shooting precision and  detection precision.
The shooting precision was calculated as the number of attacks that
did damage to an enemy (by a direct hit, a blast from an exploding
rocket, or exploding barrel) divided by the number of all performed
attacks. The detection precision is the number of attacks performed
when another player was visible to the agent divided by the number
of all performed attacks. The engine also counted the number of hits
and damage taken (in game's health points) by the agents and the number
of picked up items. The statistics are presented in Tables~\ref{tab:add-stats-track1}~and~\ref{tab:add-stats-track2}.

\paragraph{Track 1}

The level of the submitted bots was significantly higher in 2017 than
in 2016. There were no weak bots in Track 1. The spread of the frag
count was rather small: the worst bot scored 109 while the best one
248. The track was won by Marvin, which scored 248 frags, only 3 more
than the runner-up, Arnold2, and 33 more then Axon. Interestingly,
Marvin did not stand out with his accuracy or ability to avoid rockets;
it focused on gathering resources: medkits and armors, which greatly
increased his chances of survival. Marvin was hit the largest number
of times but, at the same time, it was killed the least frequently.
Arnold2, on the other hand, was better at aiming (shooting and detection
precision). 

Notice also that F1, the winner of Track 1 of the previous competition,
took the fifth place with 164 frags and is again characterized by
the least number of suicides, which, in general, did not decrease
compared to the 2016 competition and is still high for all the agents.

\begin{table}
\caption{\label{tab:results-track1-2017}Results of the 2017 Competition: Track
1. `F/D' denotes Frags/Death. Deaths include suicides.}

\centering{}%
\begin{tabular*}{1\linewidth}{@{\extracolsep{\fill}}ccccccc}
\toprule 
Place & Bot & Frags & F/D ratio & Kills & Suicides & Deaths\tabularnewline
\midrule
1 & Marvin & \textbf{248} & \textbf{1.16} & \textbf{315} & 67 & \textbf{213}\tabularnewline
2 & Arnold2 & 245 & 0.84 & 314 & 69 & 291\tabularnewline
3 & Axon & 215 & 0.77 & 252 & 37 & 278\tabularnewline
4 & TBoy & 198 & 0.60 & 229 & 31 & \emph{330}\tabularnewline
5 & F1 & 164 & 0.57 & 179 & \textbf{15} & 290\tabularnewline
6 & YanShi & 158 & 0.58 & 246 & \emph{88} & 273\tabularnewline
7 & DoomNet & 139 & 0.50 & 179 & 40 & 280\tabularnewline
8 & Turmio & 132 & 0.47 & 209 & 77 & 280\tabularnewline
9 & AlphaDoom & \emph{109} & \emph{0.39} & \emph{139} & 30 & 281\tabularnewline
\bottomrule
\end{tabular*}
\end{table}

\begin{table*}
\caption{\label{tab:results-track2-2017}Results of the 2017 Competition: Track
2. `M' denotes map and `T' denotes a total statistic.}
\setlength\tabcolsep{4pt}
\resizebox{\textwidth}{!}{%
\begin{centering}
\begin{tabular}{ccccccccccccccccccccccccc}
\toprule 
\multirow{2}{*}{Place} & \multirow{2}{*}{Bot} & \multirow{2}{*}{Total Frags} & \multirow{2}{*}{F/D ratio} &  & \multicolumn{6}{c}{Kills} &  & \multicolumn{5}{c}{Suicides} &  &  & \multicolumn{5}{c}{Deaths} & \tabularnewline
\cmidrule{6-25} 
 &  &  &  &  & M1 & M2 & M3 & M4 & M5 & T &  & M1 & M2 & M3 & M4 & M5 & T &  & M1 & M2 & M3 & M4 & M5 & T\tabularnewline
\midrule
1 & Arnold4 & \textbf{275} & 1.25 &  & 35 & 49 & 50 & \textbf{84} & 57 & \textbf{275} &  & \textbf{0} & \textbf{0} & \textbf{0} & 0 & \textbf{0} & \textbf{0} &  & 36 & \emph{48} & 42 & 44 & 50 & 220\tabularnewline
2 & YanShi & 273 & \textbf{1.47} &  & \textbf{56} & \textbf{67} & \textbf{55} & 56 & 41 & \textbf{275} &  & 1 & 1 & \textbf{0} & 0 & \textbf{0} & 2 &  & \textbf{28} & \textbf{33} & \textbf{39} & \textbf{40} & 46 & \textbf{186}\tabularnewline
3 & IntelAct & 221 & 0.89 &  & 53 & 46 & 51 & 51 & 39 & 241 &  & \emph{10} & 1 & 6 & 0 & 3 & 20 &  & 36 & \emph{48} & \emph{53} & 48 & \emph{62} & 247\tabularnewline
4 & Marvin & 193 & 0.99 &  & 36 & 38 & 52 & 46 & 44 & 216 &  & 4 & 3 & 4 & 0 & \emph{12} & \emph{23} &  & 34 & 44 & \textbf{39} & 41 & 47 & 195\tabularnewline
5 & Turmio & 164 & 0.82 &  & 12 & 58 & 26 & 45 & 50 & 181 &  & 5 & 2 & \emph{7} & 0 & 3 & 17 &  & 46 & 40 & \textbf{39} & 41 & \textbf{44} & 200\tabularnewline
6 & TBoy & 139 & 0.58 &  & 26 & \emph{16} & 33 & \emph{13} & \textbf{58} & 146 &  & 7 & \textbf{0} & \textbf{0} & 0 & \textbf{0} & \emph{7} &  & \emph{50} & \emph{48} & 46 & 49 & 47 & \emph{240}\tabularnewline
7 & DoomNet & \emph{62} & \emph{0.28} &  & \emph{7} & 20 & \emph{9} & 19 & 29 & \emph{84} &  & 4 & \emph{8} & 6 & 0 & 4 & 22 &  & 36 & 38 & 42 & \emph{51} & 54 & 221\tabularnewline
\bottomrule
\end{tabular}
\par\end{centering}
}
\end{table*}

\paragraph{Track 2}

The level of bots improved also in Track 2. All of the bots scored
more than 50 frags, which means that all could move, aim, and shoot
opponents. Similarly to the result of Track 1, the gap between the
first two bots was tiny. The competition was won by Arnold4, who scored
275 frags and was closely followed up by YanShi, who scored 273.

Arnold4 was the most accurate bot in the whole competition and the
only bot that did not commit any suicide. This turned out to be crucial
to win against YanShi, who had the same number of 275 kills but committed
two suicides. YanShi, however, achieved the highest frags/death ratio
by being the best at avoiding being killed and had the highest detection
precision. These two were definitely the best compared to the other
agents. The next runner-up, IntelAct, the winner of Track 2 in the
previous competition, scored substantially fewer, 221 frags. Fewer
items on the maps in Track 2 possibly contributed to the lower position
of Marvin, which ended up in the fourth place with 193 frags.

\begin{table*}
\caption{\label{tab:add-stats-track1}Additional Statistics for the 2017 Competition:
Track 1}
\setlength\tabcolsep{6pt}
\resizebox{\textwidth}{!}{%
\begin{centering}
\begin{tabular}{ccccccccccc}
\toprule 
Place & Bot & Avg. sp. (km/h) & Attacks & Shooting Precision (\%) & Detection Precision (\%) & Hits taken & Dmg. taken (hp) & Ammo & Midkits & Armors\tabularnewline
\midrule
1 & Marvin & 37.7 & 1654 & 23.5 & 51.6 & \emph{1266} & \textbf{34445} & 164 & \textbf{354} & \textbf{272}\tabularnewline
2 & Arnold2 & 40.5 & 1148 & \textbf{32.8} & \textbf{64.3} & 674 & 46681 & 239 & 216 & 35\tabularnewline
3 & Axon & 28.9 & 918 & 27.3 & 47.9 & 556 & 42616 & 120 & 125 & 11\tabularnewline
4 & TBoy & 17.6 & \textbf{1901} & \emph{13.9} & \emph{35.4} & 637 & \emph{51206} & \emph{16} & \emph{39} & 8\tabularnewline
5 & F1 & 25.3 & \emph{587} & 29.5 & 49.8 & 583 & 46407 & 124 & 113 & 9\tabularnewline
6 & YanShi & \emph{9.7} & 1404 & 22.6 & 41.3 & \textbf{536} & 41572 & 32 & 41 & \emph{2}\tabularnewline
7 & DoomNet & \textbf{42.9} & 690 & 29.7 & 63.8 & 642 & 44820 & \textbf{873} & 163 & 57\tabularnewline
8 & Turmio & 23.0 & 928 & 27.8 & 56.5 & 577 & 43066 & 142 & 87 & 5\tabularnewline
9 & AlphaDoom & 25.9 & 971 & 17.5 & 53.2 & 672 & 43118 & 126 & 104 & 49\tabularnewline
\bottomrule
\end{tabular}
\par\end{centering}
}
\end{table*}

\begin{table*}
\caption{\label{tab:add-stats-track2}Additional Statistics for the 2017 Competition:
Track 2. `M' denotes map and `T' denotes a total statistic.}

\setlength\tabcolsep{2pt}
\resizebox{\textwidth}{!}{%
\begin{centering}
\begin{tabular}{cccccccccccccccccccccccccccccc}
\toprule 
\multirow{2}{*}{Place} & \multirow{2}{*}{Bot} &  & \multicolumn{6}{c}{Avg. speed (km/h)} &  & \multicolumn{6}{c}{Attacks} &  & \multicolumn{6}{c}{Shooting Precision (\%)} &  & \multicolumn{6}{c}{Detection Precision (\%)}\tabularnewline
\cmidrule{4-30} 
 &  &  & M1 & M2 & M3 & M4 & M5 & T &  & M1 & M2 & M3 & M4 & M5 & T &  & M1 & M2 & M3 & M4 & M5 & T &  & M1 & M2 & M3 & M4 & M5 & T\tabularnewline
\midrule
1 & Arnold4 &  & 34.7 & 28.9 & 32.0 & 30.0 & 25.4 & 30.2 &  & 495 & 449 & 544 & 660 & 258 & \emph{2403} &  & 24 & 26.9 & \textbf{31.6} & 36.5 & 38.4 & \textbf{31.3} &  & 51.7 & 53.9 & 59.4 & 58.3 & 47.9 & 55.2\tabularnewline
2 & YanShi &  & 28.0 & 25.1 & 24.8 & 26.8 & 22.7 & 25.7 &  & 1039 & \textbf{1324} & \textbf{897} & 1767 & \textbf{745} & \textbf{5772} &  & \textbf{26.8} & 21.9 & 28.2 & 26.7 & 30.2 & 26.3 &  & \textbf{79.8} & \textbf{75.6} & \textbf{77.8} & \textbf{76.1} & \textbf{72.2} & \textbf{76.4}\tabularnewline
3 & IntelAct &  & \textbf{36.0} & 26.2 & 29.6 & 31.2 & \textbf{28.0} & 30.8 &  & 536 & 346 & 490 & 1128 & 323 & 2823 &  & 26.1 & 29.5 & 31.2 & 32.9 & 31.3 & 30.7 &  & 54.6 & 57.1 & 58.2 & 68.2 & 46.6 & 60.0\tabularnewline
4 & Marvin &  & 32.8 & \textbf{32.1} & \textbf{32.6} & \textbf{34.9} & \textbf{28.0} & \textbf{32.0} &  & \textbf{1323} & 475 & 545 & 1232 & 307 & 3882 &  & 9.7 & 17.9 & 29.4 & 19.7 & 26.7 & 18.0 &  & 33.6 & 39.1 & 54.1 & 44.6 & \emph{40.3} & 41.1\tabularnewline
5 & Turmio &  & \emph{4.7} & 19.2 & \emph{11.6} & 14.9 & \emph{9.9} & \emph{11.9} &  & 873 & \emph{267} & 512 & \emph{605} & \emph{187} & 2444 &  & \emph{2.1} & \textbf{37.5} & \emph{12.1} & \textbf{39.4} & 45.5 & 20.6 &  & \emph{8.0} & 52.1 & \emph{29.9} & 76.0 & 43.6 & \emph{37.0}\tabularnewline
6 & TBoy &  & 17.0 & 17.6 & 16.1 & \emph{2.52} & 15.3 & 13.3 &  & 459 & 910 & 658 & 1618 & 197 & 3842 &  & 15.0 & 10.3 & 20.1 & \emph{8.2} & \textbf{49.7} & 13.6 &  & 49.3 & 51.4 & 52.8 & \emph{36.5} & 53.1 & 45.2\tabularnewline
7 & DoomNet &  & 13.5 & \emph{16.0} & 18.2 & 12.6 & 25.9 & 17.3 &  & \emph{161} & 409 & \emph{323} & \textbf{2006} & 159 & 3058 &  & 7.5 & \emph{8.6} & \emph{12.1} & 8.4 & \emph{23.9} & \emph{9.5} &  & 34.8 & \emph{33.7} & 59.7 & 38.7 & 45.6 & 40.4\tabularnewline
\midrule
\multirow{2}{*}{Place} & \multirow{2}{*}{Bot} &  & \multicolumn{6}{c}{Hits taken} &  & \multicolumn{6}{c}{Dmg. taken (hp)} &  & \multicolumn{6}{c}{Ammo \& Weapons} &  & \multicolumn{6}{c}{Medikits \& Armors}\tabularnewline
\cmidrule{4-30} 
 &  &  & M1 & M2 & M3 & M4 & M5 & T &  & M1 & M2 & M3 & M4 & M5 & T &  & M1 & M2 & M3 & M4 & M5 & T &  & M1 & M2 & M3 & M4 & M5 & T\tabularnewline
\cmidrule{4-30} 
1 & Arnold4 &  & 357 & 462 & 419 & 463 & \emph{501} & 2202 &  & 3838 & 5521 & 4558 & 4635 & 5318 & 23870 &  & 143 & 255 & 68 & 207 & 138 & 811 &  & \emph{0} & 7 & 4 & 0 & 2 & 13\tabularnewline
2 & YanShi &  & \textbf{282} & 370 & 417 & \textbf{433} & 487  & 1989 &  & \textbf{2984} & \textbf{3957} & \textbf{4230} & \textbf{4285} & 4899 & \textbf{20355} &  & 119 & 309 & 64 & 229 & 144 & 865 &  & \emph{0} & 12 & 5 & 0 & 6 & 23\tabularnewline
3 & IntelAct &  & 290 & \emph{504} & \emph{497} & 533 & 576 & \emph{2400} &  & 4362 & \emph{5533} & \emph{6171} & 5225 & \emph{7011} & \emph{28302} &  & \textbf{172} & 278 & \textbf{79} & \textbf{234} & \textbf{166} & \textbf{929} &  & \textbf{2} & \textbf{13} & 6 & 0 & 5 & 26\tabularnewline
4 & Marvin &  & 309 & 415 & 349 & 440 & \textbf{324} & \textbf{1837} &  & 3818 & 5051 & 4431 & 4405 & 4757 & 22462 &  & 117 & 268 & 62 & 247 & \emph{85} & 779 &  & \emph{0} & 11 & 10 & 0 & 9 & 30\tabularnewline
5 & Turmio &  & 313 & 353 & \textbf{343} & 438 & 425 & 1872 &  & 3939 & 4408 & 4373 & 4325 & \textbf{4845} & 21890 &  & \emph{46} & \emph{159} & \emph{58} & 156 & 88 & \emph{507} &  & \emph{0} & \emph{4} & \emph{2} & 0 & \emph{0} & \emph{6}\tabularnewline
6 & TBoy &  & \emph{394} & 488 & 480 & 504 & 480 & 2346 &  & \emph{5901} & 5390 & 5058 & 4965 & 5513 & 26827 &  & 141 & \textbf{365} & 67 & \emph{75} & 134 & 782 &  & 1 & 12 & \textbf{11} & 0 & \textbf{16} & \textbf{40}\tabularnewline
7 & DoomNet &  & 286 & \textbf{334} & 368 & \emph{542} & 493 & 2023 &  & 4167 & 4458 & 4641 & \emph{5375} & 6309 & 24950 &  & 99 & 235 & 73 & 125 & 141 & 673 &  & \emph{0} & 8 & 6 & 0 & 5 & 19\tabularnewline
\bottomrule
\end{tabular}
\par\end{centering}
}
\end{table*}

\begin{table}
\caption{\label{tab:frameworks_2017}Frameworks and Algorithms used in 2017
Competition}

\begin{tabular*}{1\linewidth}{@{\extracolsep{\fill}}>{\centering}p{0.2\linewidth}>{\centering}p{0.3\linewidth}>{\centering}p{0.4\linewidth}}
\toprule 
Bot & Framework used & Algorithm\tabularnewline
\midrule
Marvin & Tensorflow & A3C, learning from human demonstration\tabularnewline
Arnold2 \& Arnold4 & PyTorch & DQN, DRQN\tabularnewline
Axon & Tensorflow & A3C\tabularnewline
YanShi(Track1) & Tensorflow + Tensorpack & DFP + SLAM + MCTS + manually specified rules\tabularnewline
YanShi(Track2) & Tensorflow + Tensorpack & A3C + SLAM + MCTS + manually specified rules\tabularnewline
Turmio & Tensorflow & A3C + Haar Detector\tabularnewline
TBoy & Tensorflow & A3C + random-grouped-curriculum-learning\tabularnewline
DoomNet & PyTorch & AAC\tabularnewline
AlphaDoom & MXNet & A3C\tabularnewline
\bottomrule
\end{tabular*}
\end{table}

\subsubsection{Notable Solutions}

\begin{itemize}
\item \textbf{Marvin} (Ben Bell) won Track 1 and took the fourth place in
Track 2. It was a version of the A3C algorithm, pre-trained with  replays
of human games collected by the authors of the bot and subsequently
trained with traditional self-play against the built-in bots. The
policy network had separate outputs for actions related to moving
and aiming, which also included aiming at the vertical axis. Additionally,
a separate policy for aiming was handcrafted to overwrite the network's
decisions if it was unsure of the correct aiming action.
\item \textbf{Arnold2 \& Arnold4} (Guillaume Lample, Devendra Singh Chaplot)
-- slightly modified versions of the 2016 runner-up Arnold; they
differ from the original mostly by the lack of separate DQN network
for navigation, support for strafing and disabled crouching. The latter
two changes might explain agent's progress -- proper strafing makes
the game considerably easier and the lack of crouching encourages
to develop more globally useful behaviors. To address agent's low
mobility, which had been originally alleviated by a separate navigation
network, Arnold 2 and 4 were hardcoded to run straight after being
stationary for too long (about 1 second). Both Arnold 2 and 4 use
the same approach and differ only in track-specific details (e.g.,
manual weapon selection).  
\item \textbf{Axon} (Cheng Ge; Qiu Lu Zhang; Yu Zheng) -- the bot that
took the third place in Track 1 was trained using the the A3C algorithm
in a few steps: first the policy network was trained on a small dataset
generated by human players, then it was trained on various small tasks
to learn specific skills like navigation or aiming. Finally, the agent
competed against the F1 agent on different maps. The policy network
utilized five scales: i) original image, ii) three images covering
middle parts of the screen and iii) one image zoomed on the crosshair.
\item \textbf{YanShi} (Dong Yan, Shiyu Huang, Chongxuan Li, Yichi Zhou)
took the second place in Track 2. Their bot explicitly separated the
perception and planning problems. It consisted of two modules. The
first one combined Region Proposal Network (RPN)~\cite{DBLP:journals/corr/HuangR17a},
which detects resources and enemies and was trained with additional
supervision using labeled images from the ViZDoom engine. The network
was combined with a Simultaneous Localization And Mapping (SLAM) algorithm,
Monte Carlo Tree Search (MCTS), and a set of hardcoded rules. The
output from the first part was combined with the second part that
utilized the code of the agents of the year 2016: F1 agent (for Track
1) and IntelAct (for Track 2). Like Marvin, YanShi handles vertical
aiming by implementing an aimbot based on the output of the RPN network.
Manually specified set of rules contains fast rotation using mouse
movement to scan the environment, dodge and prevent getting stuck
in the corner of the map. 
\end{itemize}

\subsubsection{Discussion}

The average level of the agents was arguably higher in the 2017 competition
than in the previous year. This is evidenced by the fact that the
agents from the previous competition (F1, IntelAct) were defeated
by the new submissions. Surprisingly, the largest improvement is visible
in Track 1, where the new champion scored 50\% more points than the
former one, who ended up on the fifth place. Nevertheless, the bots
are still weaker than humans especially on much harder Track 2. One
of the authors of the paper can consistently defeat all of them by
a considerable margin, although it requires some effort and focus..

In the 2017's competition, there were several important advancements
such as agents capable of effective strafing and vertical aiming.
Nonetheless, agents did not exhibit more sophisticated tactics such
as aiming at legs (much higher chance of blast damage), which is an
obvious and popular technique for human players. A3C was the method
of choice for RL while learning from demonstration (Marvin, Axon)
has started to become a common practice to bootstrap learning and
address the problem of sparse, delayed rewards. 

In addition to the algorithmic improvements, the synchronized multi-player
support in ViZDoom 1.1 allowed faster training for the 2017's competition.
New features and availability of the winning solutions from the previous
competition also opened new possibilities, allowing for a broader
range of supervised methods and training agents against other solutions
(YanShi), not only the built-in bots.

Disappointingly, bots committed a similar number of suicides in 2017
as in 2016 (Track 1). This is directly connected to the low precision
of the performed attacks and inability to understand the surroundings.
As a result, the agents often shoot at walls, wounding, and often
killing themselves. While human players have varying shooting precision,
their detection precision is usually close to 100\% (i.e., they do
not fire when they do not see the enemy). For most agents, the detection
precision decreases on the unknown maps of Track 2 and varies significantly
depending on the type of environment. It was observed that, for example,
specific textures caused some (apparently, untrained) bots to fire
madly at walls. 

Due to small map sizes, the agents encountered each other often. It
was also noticed that the agents have little (if any) memory ---
they often ignore and just pass by each other, without trying to chase
the enemy, which would be natural for human players.

\subsubsection{Logistics}

In the 2017 competition, the solutions were submitted in the form
of Docker images, which made the preparation of software environments
easier, removed most of the compatibility issues and unified the evaluation
procedure. Nevertheless, the need for manual building and troubleshooting
some of the submissions remained. This has shown that there is a need
for a more automated process, preferably one where solutions can be
submitted on a daily basis and are automatically verified and tested
giving immediate feedback to the participants.

\section{Conclusions\label{sec:Conclusions}}

This paper presented the first two editions of Visual Doom AI Competition,
held in 2016 and 2017. The challenge was to create bots that compete
in a multi-player deathmatch in a first-person shooter (FPS) game,
Doom. The bots were to act based on the raw pixel information only.
The contests got a large media coverage\footnote{e.g., \url{https://www.theverge.com/2016/4/22/11486164/ai-visual-doom-competition-cig-2016}
or \url{https://www.engadget.com/2016/08/18/vizdoom-research-framework-cig-competition/}} and attracted teams from leading AI labs around the world.

The winning bots used a variety of state-of-the-art (at that time)
RL algorithms (e.g., A3C, DRQN, DFP). The bots submitted in 2017 got
stronger by fine-tuning the algorithms and using more supervision
(human-replays, curriculum learning). It was also common to separately
learn to navigate and to fight. The bots are definitely competent
but still worse than human players, who can easily exploit their weaknesses.
Thus, a deathmatch from a raw visual input in this FPS game remains
an open problem with a lot of research opportunities. 

Let us also notice that the deathmatch scenario is a relatively easy
problem compared to a task of going through the original single-player
Doom levels. It would involve not only appropriate reaction for the
current situation but also localization and navigation skills on considerably
more complex maps with numerous switches and appropriate keys for
different kinds of doors, which need to be found to progress. Therefore,
the AI for FPS games using the raw visual input is yet to be solved
and we predict that the Visual Doom AI competition will remain a difficult
challenge in the nearest future. To further motivate research towards
solving this challenging problem, in the upcoming edition of the competition
(2018), the form of Track 1 has been changed. The new task is to develop
bots that are capable of finishing randomly generated single-player
levels, ranging from trivial to sophisticated ones, that contain all
the elements of the original game.

The paper also revisited ViZDoom (version 1.1.5), a Doom-based platform
for research in vision-based RL that was used for the competitions.
The framework is easy-to-use, highly flexible, multi-platform, lightweight,
and efficient. In contrast to other popular visual learning environments
such as Atari 2600, ViZDoom provides a 3D, semi-realistic, first-person
perspective virtual world. ViZDoom's API gives the user full control
of the environment. Multiple modes of operation facilitate experimentation
with different learning paradigms such as RL, adversarial training,
apprenticeship learning, learning by demonstration, and, even the
`ordinary', supervised learning. The strength and versatility of the
environment lie in its customizability via a mechanism of scenarios,
which can be conveniently programmed with open-source tools. The utility
of ViZDoom for research has been proven by a large body of research
for which it has been used (e.g.,~\cite{Chaplot2016TransferDR,DBLP:journals/corr/ChaplotSPRS17,DBLP:journals/corr/PathakAED17,DBLP:journals/corr/BhattiDMNST16,DBLP:journals/corr/ParisottoS17}).

\section*{Acknowledgment }

This work has been supported in part by Ministry of Science and Education
grant no. 91--500/DS. M. Kempka acknowledges the support of Ministry
of Science and Higher Education grant no. 09/91/DSPB/0602.

\bibliographystyle{plain}
\bibliography{FPS-research,bibliography,wjaskowski,all,library}

\appendices{}

\appendix{Online Resources}
\begin{enumerate}
\item ViZDoom: \url{https://github.com/mwydmuch/ViZDoom}
\item Submissions in 2016 (docker files):\\\url{https://github.com/mihahauke/VDAIC2017}
\item Submissions in 2017 (docker files):\\\url{http://www.cs.put.poznan.pl/mkempka/misc/vdaic2017}
\item Map packs used in 2017:\\\url{https://github.com/mihahauke/VDAIC2017}
\item Videos from 2016:\\\url{https://www.youtube.com/watch?v=94EPSjQH38Y}\\\url{https://www.youtube.com/watch?v=tDRdgpkleXI}\\\url{https://www.youtube.com/watch?v=Qv4esGWOg7w}
\item Videos from 2017:\\\url{https://www.youtube.com/watch?v=3VU6d_5ze8k}\\\url{https://www.youtube.com/watch?v=hNyrZ5Oo8kU}
\end{enumerate}

\end{document}